\pdfoutput=1
\documentclass{edm_article}

\usepackage{booktabs}
\usepackage{algorithm}
\usepackage[noend]{algpseudocode}
\usepackage{graphicx}
\usepackage{tabularx}
\usepackage{textcomp}
\usepackage{xcolor}
\usepackage{enumitem}

\usepackage{color}
\usepackage{subcaption}
\definecolor{pblue}{rgb}{0.13,0.13,1}
\definecolor{pgreen}{rgb}{0,0.5,0}
\definecolor{pred}{rgb}{0.9,0,0}
\definecolor{pgrey}{rgb}{0.46,0.45,0.48}

\usepackage{listings}
\begin{document}

\title{Generative Grading: Near Human-level Accuracy for Automated Feedback on Richly Structured Problems}


\author{
Ali Malik$^{1\textbf{*}}$, Mike Wu$^{1\textbf{*}}$,\\ Vrinda Vasavada$^1$, Jinpeng Song$^1$, Madison Coots$^1$,\\ John Mitchell$^1$, Noah Goodman$^{1,2}$,  Chris Piech$^1$ \\ 
\\
{$^1$\affaddr{Department of Computer Science, Stanford University}} \\
{$^2$\affaddr{Department of Psychology, Stanford University}} \\
\email{\footnotesize \{malikali, wumike, vrindav, jsong5, mcoots, jcm, ngoodman, piech\}@cs.stanford.edu}
}

\maketitle


\begin{abstract}

Access to high-quality education at scale is limited by the difficulty of providing student feedback on open-ended assignments in structured domains like computer programming, graphics, and short response questions. This problem has proven to be exceptionally difficult: for humans, it requires large amounts of manual work, and for computers, until recently, achieving anything near human-level accuracy has been unattainable. In this paper, we present generative grading: a novel computational approach for providing feedback at scale that is capable of accurately grading student work and providing nuanced, interpretable feedback. Our approach uses generative descriptions of student cognition, written as probabilistic programs, to synthesise millions of labelled example solutions to a problem; we then learn to infer feedback for real student solutions based on this cognitive model.

We apply our methods to three settings. In block-based coding, we achieve a 50\% improvement upon the previous best results for feedback, achieving super-human accuracy. In two other widely different domains---graphical tasks and short text answers---we achieve major improvement over the previous state of the art by about 4x and 1.5x respectively, approaching human accuracy.  In a real classroom, we ran an experiment where we used our system to augment human graders, yielding doubled grading accuracy while halving grading time.

\end{abstract}

%


\section{Introduction}
Enabling global access to high-quality education is a long-standing challenge.
The combined effect of increasing costs per student \cite{bowen2012cost} and rising demand for higher education makes this issue particularly pressing. A major barrier to providing quality education has been the ability to automatically provide \emph{meaningful and accurate} feedback on student work, particularly for open-ended tasks like programming or simple natural language response.





\begin{figure}[t]
    \centering
    \includegraphics[width=0.9\linewidth]{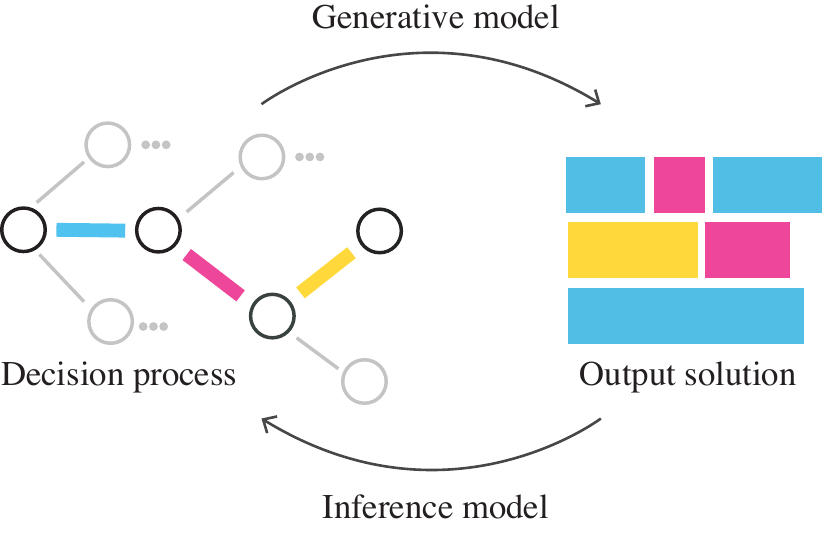}
    \caption{\small Students solve problems by making decisions that result in their final solution (generative model). Providing feedback requires the reverse task of seeing a solution and inferring the student decisions that lead to this solution (inference model).}
    \label{fig:overview}
\end{figure}

Learning to provide feedback on richly structured problems beyond simple multiple-choice has proven to be a hard machine learning problem. Five issues have emerged, many of which are typical of human-centred AI problems: (1) student work is extremely diverse, exhibiting a heavy tailed distribution where most solutions are rare, (2) student work is difficult and expensive to label with fine-grained feedback, (3) we want to provide feedback (without historical data) for even the very first student, (4) grading is a precision-critical domain with a high cost to misgrading, and (5) predictions must be explainable and justifiable to instructors and students.  Despite extensive research that combines massive education data with cutting-edge deep learning  \cite{piech2015deep,basu2013powergrading,yan_pyramid,wang2017learning,liu2019automated,hu2019reliable}, these issues make traditional supervised learning inadequate for automatic feedback at scale.

Human instructors are experts at providing student feedback. When grading assignments, they have an understanding of the decisions and missteps students might make when solving a problem, and what corresponding solutions these choices would result in.
For example, an instructor understands that a student wanting to repeat something in a programming assignment might use a \texttt{for} loop or manually write repeated statements. And given that the student uses a loop, their loop could be correct or off-by-one.

In essence, instructors mentally possess \textit{generative models} of student decision-making and how these decisions manifest in a final solution (Fig. \ref{fig:overview}, forwards). When providing feedback, the instructor does \textit{inference}: given a student solution, they use their mental model to try and determine the underlying student decisions that could have resulted in this solution (Fig. \ref{fig:overview}, backwards).

In this paper, we propose an automated feedback system that mimics the instructor process as closely as possible. Firstly, the system elicits a literal generative model from an instructor in the form of a concrete student simulator. Secondly, it uses deep neural networks and a novel inference method with this simulator to learn how to do inference on student solutions, \textit{without using any labelled data}. Finally, the inference model is used to provide automated feedback to real student solutions. We call this end-to-end approach \textbf{generative grading}.

When used across a spectrum of public education data sets, our automated feedback system is able to grade student work with close to \textit{expert human-level} fidelity.
In block-based coding, we attain super-human accuracy, achieving a 50\% improvement over the previous best results for feedback. In two other widely different domains---graphical tasks and short text answers---we achieve major improvement over the previous state of the art by about 4x and 1.5x respectively, approaching human accuracy.
We used our system in a real classroom to augment human graders in a CS1 class, yielding \textit{doubled} grading accuracy while \textit{halving} grading time.

\subsection{Main contributions}

In Sec.~\ref{sec:generative_model}, we present an easy-to-use and highly expressive class of generative models called Idea2Text simulators that allow an instructor to encode their mental models of student decision-making. These simulators can succinctly express student decisions and how these decisions manifest in a final solution for a broad set of problem domains like graphics programming, short-answer questions, and introductory programming in Java. We provide a Python implementation that allows any instructor to easily write these simulators.

In Sec.~\ref{sec:inference}, we show how to use Idea2Text simulators with deep neural networks to infer students' decision processes from their solutions. This extracted decision process is a general representation of a student's solution and can be used for several downstream tasks such as providing automated feedback, assisting human grading, auditing and interpreting the model decisions, and improving the quality of the simulator itself.

In order to do inference successfully on our expressive class of simulators, we must overcome several interesting technical challenges (Sec.~\ref{sec:inference}). Learning to map solutions to sequences of decisions specified by the simulator is a nonstandard machine learning task, with non-fixed labels, varied sequence lengths, and unexpected trajectories. Moreover, generating simulated training data from the simulators requires an intelligent sampling method to work effectively.

In Sec.~\ref{sec:experiments} we show the efficacy of our approach in practice on a diverse set of richly structured problems. We attain close to human level accuracy on providing feedback and surpass many previous state of the art results. We also discuss several interesting extensions in Sec.~\ref{sec:extensions} that use our system to go beyond just providing automated feedback.

The generative grading system is powerful because it addresses many of the issues of traditional supervised learning mentioned above. We find that the cost of writing simulators for a new assignment is orders of magnitude cheaper for instructors than manually annotating individual student work. The simulators allow us to sample infinite data, and our adaptive sampling strategy lets us explore diverse student solutions in our training data. It is ``zero-shot", requiring no historical data nor annotation, and thus works for the very first student.  Moreover, our novel inference system allows for interpretable and explainable decisions.






\section{Related Work}
\label{sec:related}
 ``Rubric sampling'' \cite{wu2018zero} first introduced the concept of encoding expert priors in grammars of student decisions, and was the inspiration for our work. The authors design Probabilistic Context Free Grammars (PCFGs) to curate synthetically labelled datasets to train supervised classifiers for feedback. Our approach builds on this, but presents a more expressive family of generative models of student decision-making that are context sensitive and comes with new innovations that enable effective inference. From our results on Code.org, we see that this expressivity is responsible for pushing our model past human level performance. Furthermore, this prior work only used to PCFGs to create simulated datasets of feedback labels for supervised learning. In contrast, we learn to infer the entire decision trajectory of a student solution, allowing us to do things like dense feedback and human-in-the-loop grading.


We draw theoretical inspiration for our generative
grading system  from Brown’s ``Repair Theory"
which argues that the best way to help students is to understand the generative origins of their mistakes \cite{brown1980repair}. Building systems of student cognition has been used in K-12 arithmetic problems  \cite{koedinger2015methods} and subtraction mistakes \cite{feldman2018automatic}.

Automated feedback for open-ended richly structured problems has been studied through a few lenses. In many approaches, traditional supervised learning is employed to map solutions to feedback \cite{feedbackEssayNilforoshan, feedbackEssayWoods, basu2013powergrading, yan_pyramid}. These methods require large hand-labelled datasets of diverse student solutions, which is difficult due to heavy-tailed distributions. Feedback specific to computer programming problems has been explored based on executing student solutions and comparing to a reference solution \cite{gulwani2013automated,feedbackFormalSemantics}. An interesting parallel to our work is found in \cite{gulwani2013automated}, where the instructor is asked to specify the kinds of mistakes students can make. These approaches are limited to code and don't provide feedback on the problem-solving process of a student.

Extracting expert-written generative models for inference has seen enormous use in fields where domain expertise is critical. Some key example include medical diagnosis, engineering, ecology, and finance, where a generative model like a Bayesian network is elicited from experts. \cite{expertconservation, expertengineering}. In education, instructors have domain expertise about students, and Idea2Text serves as an easy-to-use generative model for instructors to encode this expertise.

Inference over decision trajectories of Idea2Text simulators is similar to ``compiled inference" for execution traces in probabilistic programs. As such, our inference engine shares similarities to this literature \cite{ritchie2016DAIPP,le2016inference}. With Idea2Text simulators, we get a nice interpretation of compiled inference as a parsing algorithm.

\section{Background}
\label{sec:background}
In this section, we introduce the feedback challenge and what makes it a difficult machine learning problem.

\subsection{Feedback as a Prediction Problem}
The \textit{feedback prediction} task is to automatically provide feedback to a given student solution. While this is easy to do for simple multiple-choice problems, we focus on the challenging task of providing feedback on richly-structured problems like computer programming or short-answer responses.

Both the type of student solutions and the type of feedback required for the task can take many forms. A student solution can be a piece of text, which could represent problems like an essay, a maths proof, or a code snippet. It could also be graphical output in the form of an image. Similarly, feedback can take the form of something simple like classifying solutions to a fixed set of misconceptions, or something complex such as highlighting and annotating specific parts of a student solution

\subsection{Difficulty of Automated Feedback}
\label{sec:difficulty}

Feedback prediction on richly structured problems has been an extremely difficult challenge in education research. Even limited to simple problems in computer science like beginner block-based programming, automated solutions to providing feedback have been restricted by scarce data and lack of robustness.
We discuss a few of the properties of student work that make predicting feedback such a difficult challenge in education.

\begin{figure}[t!]
    \centering
    \begin{subfigure}{0.24\linewidth}
        \centering
        \caption{Code.org}
        \includegraphics[width=\textwidth]{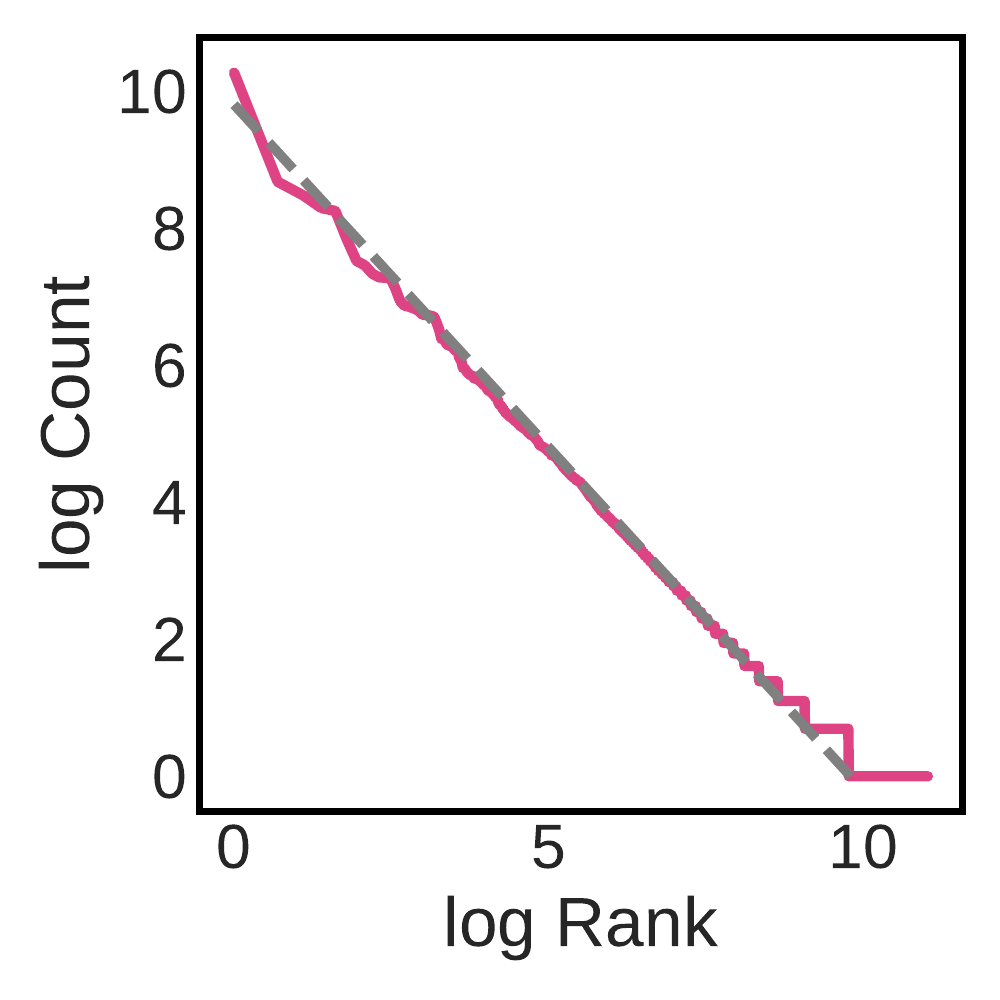}
    \end{subfigure}
    \begin{subfigure}{0.24\linewidth}
        \centering
        \caption{Liftoff}
        \includegraphics[width=\textwidth]{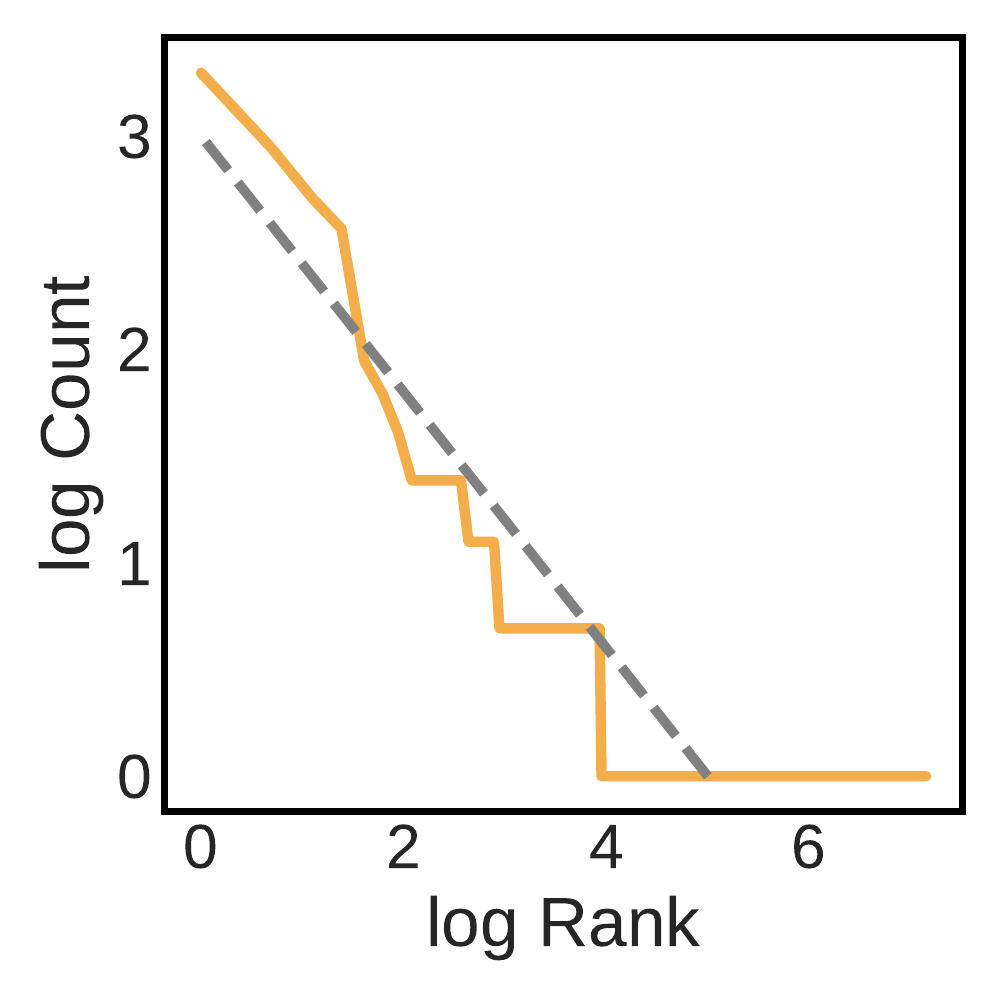}
    \end{subfigure}
    \begin{subfigure}{0.24\linewidth}
        \centering
        \caption{Pyramid}
        \includegraphics[width=\textwidth]{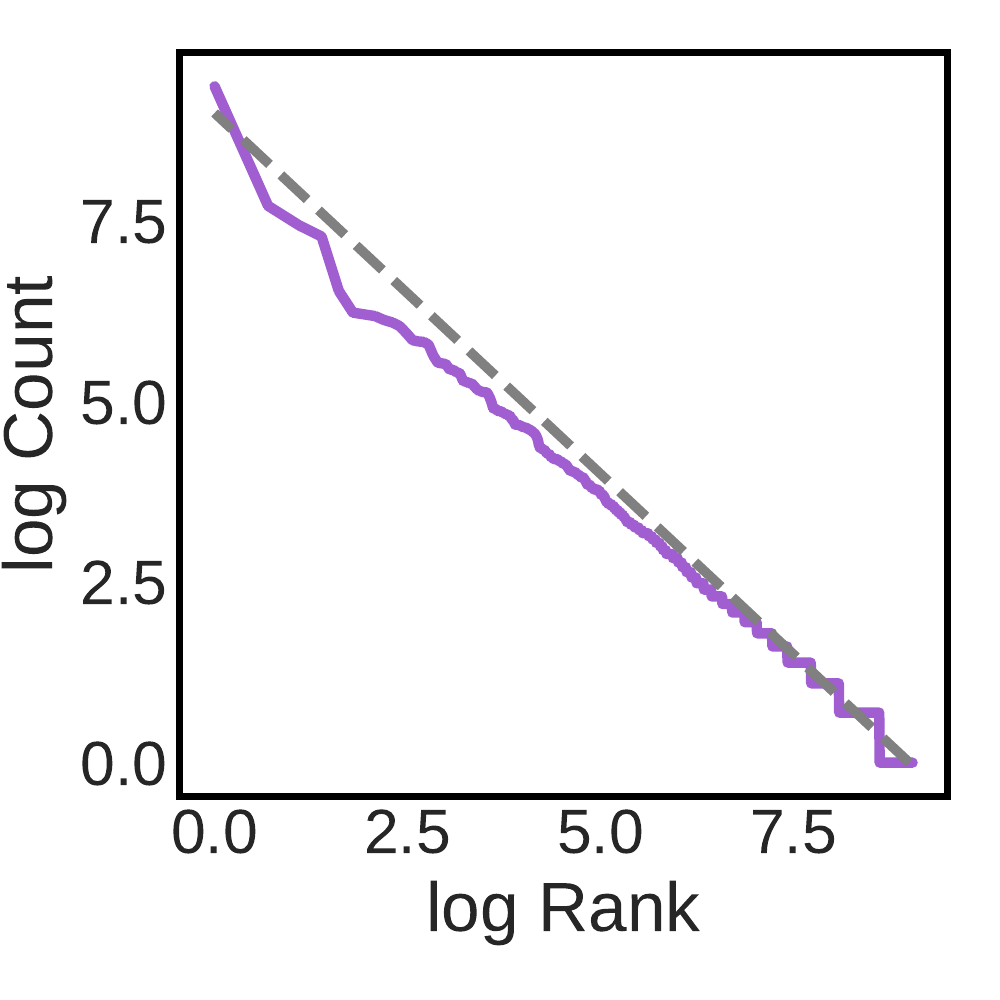}
    \end{subfigure}
    \begin{subfigure}{0.24\linewidth}
        \centering
        \caption{Power}
        \includegraphics[width=\textwidth]{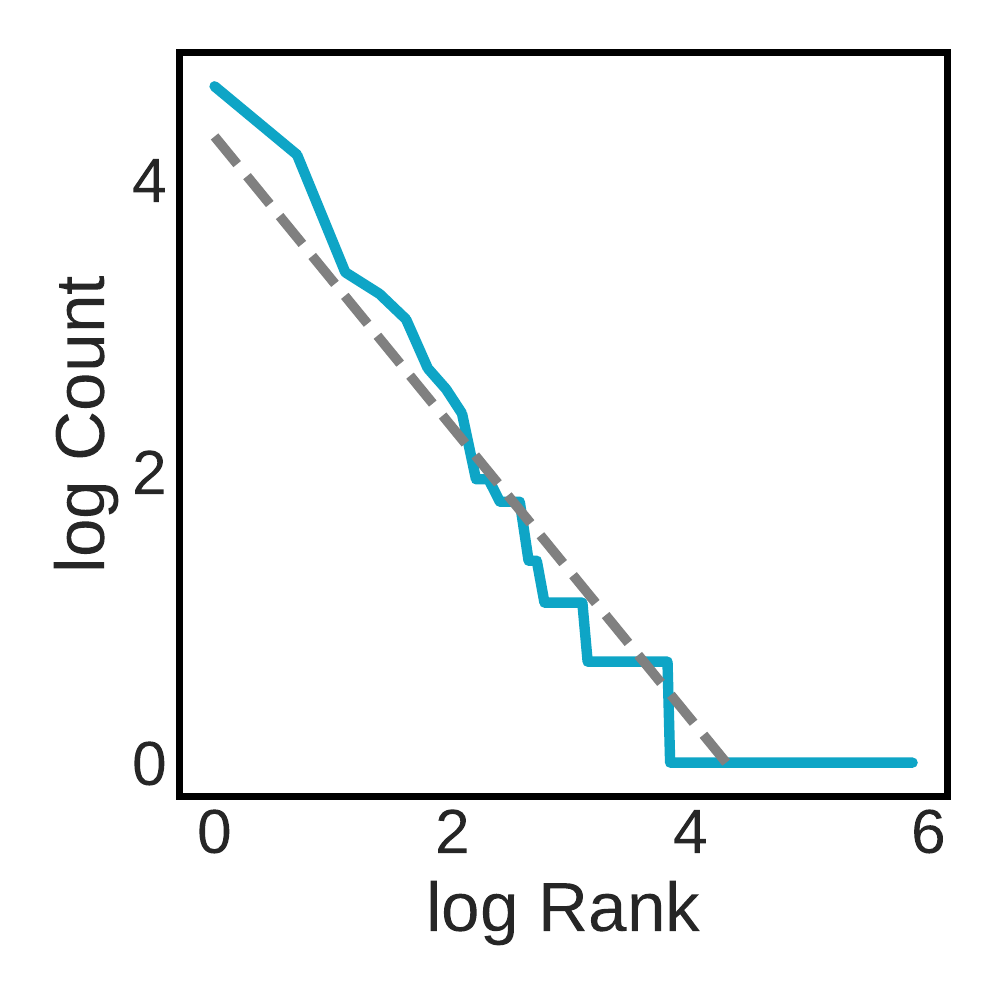}
    \end{subfigure}
    \caption{\small Student solutions (across many domains) exhibit heavy-tailed Zipf distributions, meaning a few solutions are extremely common but all other solutions are highly varied and show up rarely. This suggests that the probability of a student submission not being present in a dataset is high, making supervised learning on a small data set difficult. }
    \label{fig:zipf}
\end{figure}

\noindent\textbf{(1) Heavy-tailed Distributions:} Student work in the form of natural language, mathematical symbols, or code follow heavy-tailed Zipf distributions. This means that a few solutions are extremely common whereas almost all other examples are unique and show up rarely. Fig.~\ref{fig:zipf} plots the log-frequency of unique examples against the log of the rank across four datasets of student work in block-based programming code, Java code, and free response. For all datasets, we observe a linear relationship in log-log space, which is a characteristic property of Zipf distributions.

These heavy-tailed Zipf distributions pose a hard generalisation problem for traditional supervised machine learning: a handful of similar examples appear very frequently in the training data whereas almost all other examples are unique. This means at test time, examples are likely to introduce unseen tokens, new misconceptions, and novel student  strategies. In a Zipf distribution, even if we observe a million student solutions, there is roughly a 15\% chance that the next student generates a unique solution.

\noindent\textbf{(2) Difficulty of Annotation:} Annotating student work with feedback requires an instructor level expertise of the domain. Providing fine-grained feedback also takes effort as the annotator must read and understand student solutions before inferring possible misconceptions. In \cite{wu2018zero}, the authors found that 800 student block-based programmings solutions took 26 hours to label.

This difficulty, combined with Zipf properties, makes supervised learning intractable.
Even in the extreme cases, annotation does not scratch the surface of the Zipf distribution. As an example, in 2014, Code.org, a widely used resource for beginners in computer science, ran an initiative to crowdsource thousands of instructors to label 55,000 student solutions in a block-based programming language. Yet, despite having access to an unprecedented amount of labelled data, traditional supervised methods failed to perform well on even these ``simple'' questions.

\noindent\textbf{(3) Limitations on Data Size:} Even if the Code.org approach succeeded, most classrooms do not share the same scale as Code.org. A method that relies heavily on historical data is not widely applicable in the average classroom setting. In our experiments, our data sets contain less than a few hundred examples, again barring application of supervised algorithms. The ideal feedback model will be zero-shot so that it works even for the very first student.

\section{Modelling Student Cognition}
\label{sec:generative_model}

Having presented the feedback challenge and motivated why supervised learning cannot solve this problem alone, we discuss the idea of modelling the cognitive process of student decision-making when producing a solution.

When an instructor provides feedback on student work, they often have a latent mental model of student decision making; this captures the kinds of steps and mistakes they think students will make and what solutions are indicative of those steps. The instructor (or TAs) then grade solutions, one at a time, by essentially inferring the steps in the decision-making model that lead to the produced solution.

As a concrete example from introductory programming, suppose a student is trying to print a countdown of numbers from 10 to 1. An instructor understands that as a first step, a student might use a \texttt{for} loop or manually write ten \texttt{print} statements. Given that the student uses a loop, they could increment up or down. And given that they increment down, their loop comparison could be correct or off-by-one.  At each of these decision points, the instructor can conceive how a specific choice would manifest in the solution.

We want to allow instructors to capture their mental model of student  decision-making for a given problem and distil it into a concrete executable simulator that generates solutions. Such a generative model could be used to simulate unlimited student data, including their decision process and resulting solutions. This simulated dataset can then be used for learning how to provide feedback.

In this section, we formalise the idea of a student's  decision process for generating the solution to a problem and discuss how we can represent the instructor's latent model of this generative process as a concrete simulator.

\subsection{Student Decision Process (SDP)}

A student's decision process (SDP) can be seen as the sequence of choices the student makes while solving a problem, and how those choices correspond to different outputs.  The sequence is intended to reflect all of the critical decisions made by the student, and in particular those a teacher would like to recover from the solution. Importantly not all students will encounter the same sequence of decisions since an early decision choice may determine which decisions are faced later in problem solving.

 We can formally think of a decision point as a categorical random variable, $X$, and the specific choices that can be made as the different values, $x \in Val(X)$, that the random variable can take. The decision process of a student can be seen as a sequence trajectory $(X_t, x_t)_{t=1}^T$, of decisions encountered and made by the student in solving the problem.

Under this interpretation, specifying a model for an SDP amounts to defining the space of possible decision trajectories and a probability distribution over this space. By the chain rule for probabilities, we can decompose this overall distribution as a sequence of conditional probabilities
$\mathbb{P}(X_t = x_t \mid \mathbf{X}_{<t} = \mathbf{x}_{<t})$.
Here $x_t$ is the choice made at step $t$ from domain $X_t$ and
$\mathbf{X}_{<t}$ is shorthand for the sequence of decisions before time $t$.

We want to make this general formulation more tractable to allow us to specify useful SDPs as generative models we can sample from. Prior work \cite{wu2018zero} has attempted to express SDPs by restricting the class of generative models to probabilistic context free grammars (PCFGs).
They found that instructor-written PCFGs could often be used to emulate student problem-solving and generate student solutions for small problems.
In this setting, the non-terminal nodes of the PCFG represent decisions to be made (e.g. syntax confusion) and the production rules represent how decisions are made and manifested into output (e.g. the code is missing a semicolon). Instructors create student simulators by specifying decision points, rules, and probabilities for each rule (e.g. missing a semicolon is much more likely than missing a \texttt{function} statement).


A PCFG is compact and useful, but makes the independence assumption that that the choice made at time $t$ is independent of past choices made while solving the problem i.e. $\mathbb{P}(X_t = x_t \mid \mathbf{X}_{<t} = \mathbf{x}_{<t}) = \mathbb{P}(X_t = x_t)$.
This context-independence is a strong restriction that severely limits what instructors can express about the student decision-making process and fails to faithfully model student reasoning. As can be seen in even the simple countdown example above, the off-by-one error would manifest differently in student output depending on whether the student chose to increment up or down. Thus context dependence of decision making is an important property to model.

\begin{algorithm}[t]
    \small
    \caption{{\footnotesize Idea2Text Simulation}}
    \label{alg:idea2text}
    \textbf{Input:} Idea2Text simulator $(D, \Sigma, S)$\;

    \textbf{Output:} Tuple $(\tau, y)$ of decision trajectory and output solution.\;

    \begin{algorithmic}[1]
        \Procedure{Simulate}{$D$, $\Sigma$, $S$}
            \State $\tau \gets [ \ ]$
            \State $y \gets \Call{Generate}{S, \tau}$\Comment{Begin from start node}
            \State \Return $(\tau, y)$
        \EndProcedure

        \Procedure{Generate}{$N$, $\tau$}

            \State $a, X_a, \Pi_a \leftarrow N$\Comment{Unpack current decision node}

            \State $x_a, y \gets \Pi_a(X_a, \tau)$\Comment{Get decision choice and output}

            \State $\tau.{append}((a, x_a))$

            \For{decision node $N'$ in $y$}\Comment{In order left to right}
                \State $y' \gets \Call{Generate}{N', \tau}$
                \State $y \gets \Call{Replace}{y, N', y'}$\Comment{Replace $N'$ with $y'$}
            \EndFor
            \Return $y$
        \EndProcedure
    \end{algorithmic}
\end{algorithm}

\subsection{Idea2Text}

In this section, we define a broader class of generative models that is powerful enough to capture more complexities of expert models of student cognition.
Similar to PCFGs we structure our models around a set of non-terminal symbols that correspond to student decisions and contribute to the final output. However, drawing from work on probabilistic programs \cite{goodman2012church, goodman2014design}, we allow these choices to be made depending on previous choices.
While dependence on context leads to extremely expressive models, we will show that requiring some text to be generated at each step is enough for inference to remain tractable.
We call this class of generative models Idea2Text.\footnote{A very similar class of models, used in the very different domain of customer service, was independently named Idea2Text by scientists at Gamalon, Inc.}

Concretely, an Idea2Text simulator consists of a tuple of $(D, \Sigma, S)$ denoting a set of nonterminal decision nodes, a set of terminal nodes, and a starting root node, respectively. Intuitively, decision nodes correspond to decisions a student might make and the terminal nodes correspond to literal text tokens in the final output. Each run of the simulator also keeps a global state $\tau$ which stores the history of all decisions made during the execution (often called an \emph{execution trace} for probabilistic programs \cite{pyro, wingategoodman}).

Each decision node in $D$ is a tuple $(a, X_a, \Pi_a)$ consisting of a unique name, a random variable representing the decision choices, and a production program, which (1) specifies how this decision should be made based on the decisions made so far, and (2) produces an output solution for a given decision choice.

\begin{figure}[t!]
    \centering
    \includegraphics[width=0.85\linewidth]{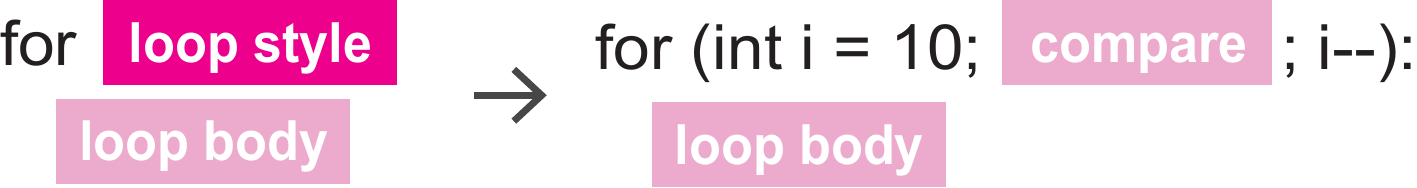}
    \caption{\small An example decision expansion step in an Idea2Text simulation. The ``loop style'' decision node chooses a type of for loop (e.g. increment vs decrement) and outputs a string containing the header of the for loop (terminals) plus another decision node.}
    \label{fig:example}
\end{figure}

More concretely, the production program is a probabilistic function that takes the current decision history $\tau$ and does the following:
\vspace{-3mm}
\begin{enumerate}[label=(\arabic*)]
    \item Samples the random variable $x_a \sim X_a$, from a distribution, $\mathbb{P}(X_a | \tau)$, that can depend on the decision history.
    \item Based on the sampled choice, $x_a$, produces an output string, $y$, consisting of literal text (terminal nodes) and incomplete segments (decision nodes). These incomplete segments correspond to future decisions that will be expanded later (see Fig.~\ref{fig:example}).
    \item Returns the sampled choice and output string: ($x_a, y$).
\end{enumerate}

An output from Idea2Text is a generation from the root node to a sequence of literal text by recursively expanding the decision nodes, as shown in Algorithm \ref{alg:idea2text}.
Each output is associated with the final decision trajectory, $\tau = [(a_t, x_{a_t})]_{t=1}^T$, of random variables encountered during generation. Here, $a_t$ denotes the unique name for the random variable encountered at timestep $t$, and $x_{a_t}$ is the sampled value for that variable.

We point out some important properties of Idea2Text simulators. First, each decision node's choice can depend on the decision history $\tau$, allowing past decisions to influence current behaviour. This is strictly more expressive than PCFGs \cite{wu2018zero} and allows instructors to write highly contextual models of student problem-solving. Second, production programs have the full power of programming languages at their disposal and can use arbitrarily complicated transformations to produce their output sequence. As an example, a production program can transform terminal nodes into images or use a machine-learning conjugator to conjugate it's produced text into a proper sentence.

\section{Inference}
\label{sec:inference}
\label{sec:models}

In this section we describe how we can use an instructor-written Idea2Text simulator to learn how to infer the decisions underlying a student solution.

At a high-level, the Idea2Text simulator contains the instructor's mental model for the sequence of decisions students make that result in different solutions. For inference we want to do the reverse: given a student solution, we want to find a trajectory of decisions in the simulator that would produce that solution.

A model that could successfully do this inference could be used to map real student solutions to decision steps in the simulator. These extracted decisions are a rich and general representation of a  student’s  solution  and  can  be  used  for  downstream  tasks  such as   automated  feedback,  assisting human grading, auditing and interpreting the model decisions, and improving the quality of the simulator.

More formally, let $\mathcal{G}$ be a given Idea2Text simulator. Each execution of $\mathcal{G}$ produces a decision trajectory $\tau$ and corresponding production $y$. Since the execution is probabilistic, the simulator induces a probability distribution $p_\mathcal{G}(\tau, y)$ over trajectories and productions.

Given a student solution, $y$, we are interested in the task of \textit{parsing}: this is the task of mapping $y$ to the most likely trajectory in the Idea2Text simulator, $\arg\max_{\tau} p_\mathcal{G}(\tau | y)$, that could have produced $y$.
This is a difficult search problem: the number of trajectories grows exponentially even for simple grammars, and common methods for parsing by dynamic programming (Viterbi, CYK) are not applicable in the presence of context-sensitivity and functional transformations.
What's more, in order to transfer robustly to real student solutions, we would like to be able to approximately parse solutions which it is not possible to generate from the simulator but are sufficiently ``nearby''.

At a high level, our approach is to construct a large data set from the simulator and then learn an inverse ``inference'' neural net that can reconstruct the decision trajectory from the solution.

\subsection{Adaptive Grammar Sampling}
To train our models, we generate a large dataset of $N$ trajectories and their associated productions, $\mathcal{D} = \{ (\tau^{(m)}, y^{(m)}) \}_{m=1}^N$, by repeatedly executing $\mathcal{G}$.









However, due to the Zipf-like nature of student work (see Sec.~\ref{sec:difficulty}), standard i.i.d. sampling from the simulator will tend to over-represent the most probable productions.
For our models, the more diverse student cognition we can simulate in the training data, the more we expect to generalise to the long tail of real students. Thus, we need sampling strategies that prioritise diversity.

A simple but flawed idea for generating diverse solutions would be to make choices at decision nodes uniformly randomly instead of using the expert-written distributions. This approach will generate more unique productions, but disregarding the expert-written distributions will result in unlikely and less realistic productions.

Ideally, we want to sample in a manner that covers all the most likely productions first, and then smoothly transition into sampling increasingly unlikely productions. This would generate unique productions efficiently while also retaining the expert-written distributions specified in the production programs.
With these desiderata in mind, we propose a method called \textit{Adaptive Grammar Sampling}. For each decision node in the simulator, we down-weight the probability of sampling each choice proportional to how many times it has been sampled in the past. To avoid overly punishing decision nodes early in the execution trace, we discount this down-weighting by a decay factor $d$ that depends on the depth of the decision in the trajectory.\footnote{The details can be found in the code.} This method is inspired by Monte-Carlo Tree Search \cite{mcts_chang} and shares similarities with Wang-Landau sampling from statistical physics \cite{wang_landau}.

\begin{figure}[t!]
    \centering
    \begin{subfigure}{0.49\linewidth}
        \centering
        \includegraphics[width=\linewidth]{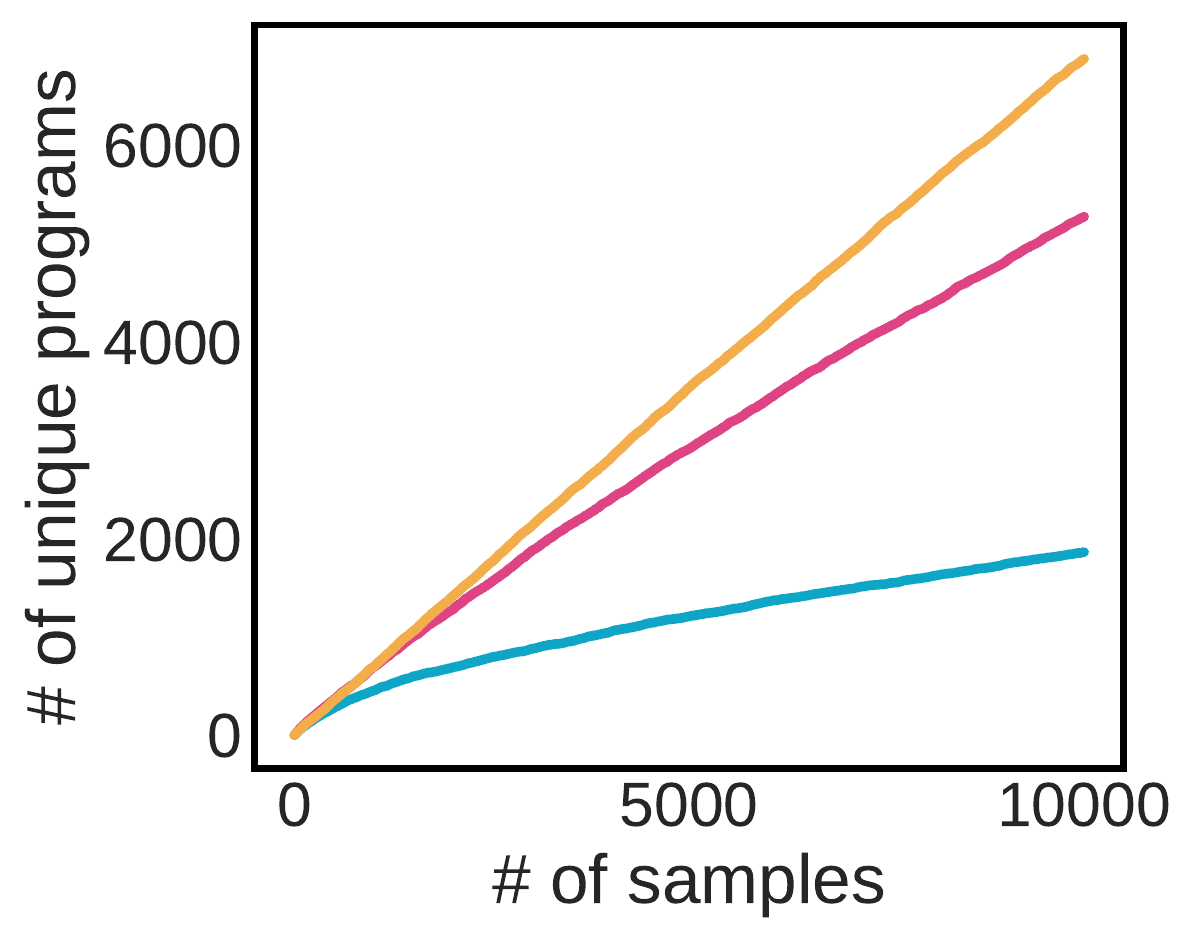}
        \caption{Uniqueness}
    \end{subfigure}
    \begin{subfigure}{0.49\linewidth}
        \centering
        \includegraphics[width=\linewidth]{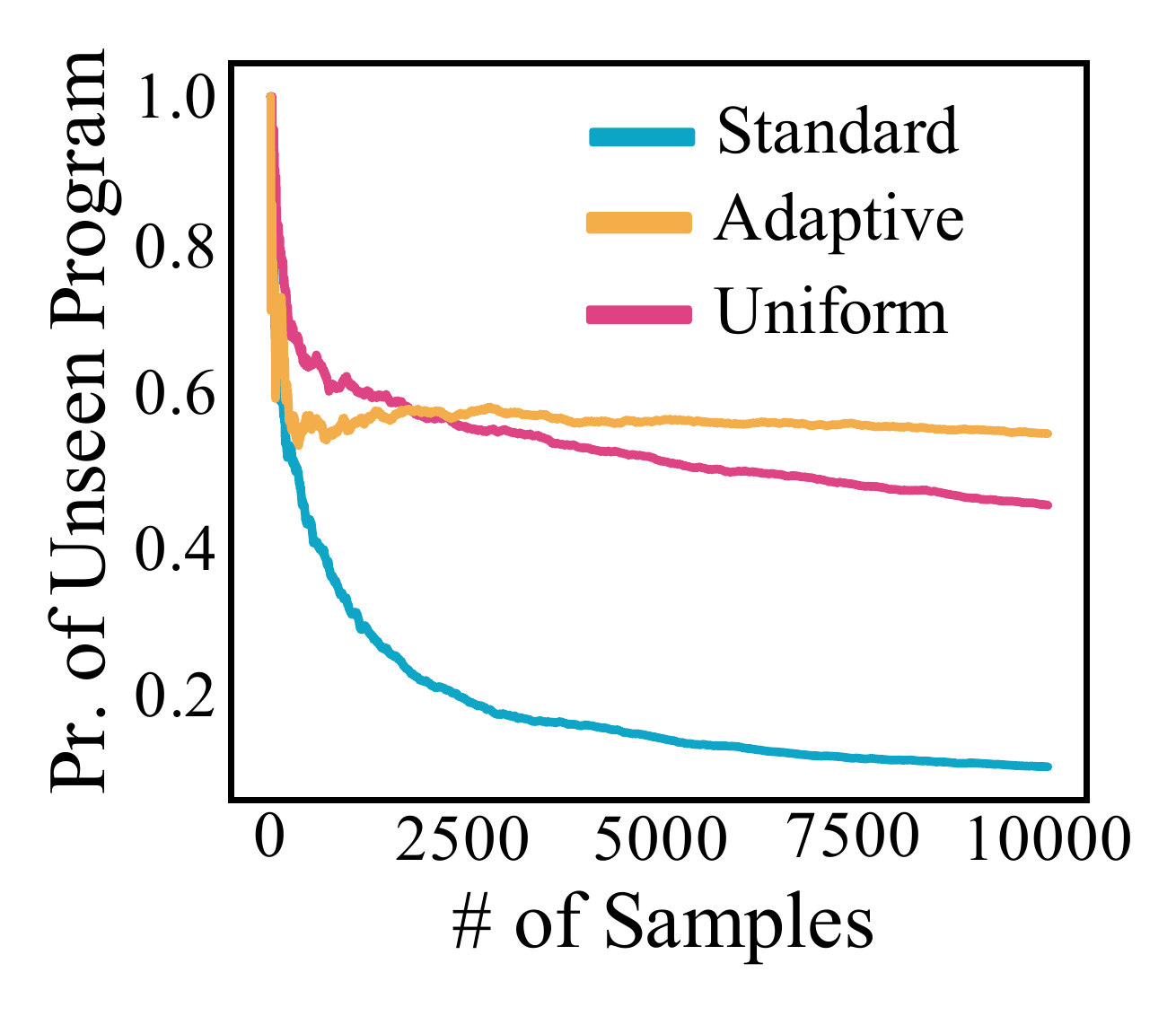}
        \caption{Good-Turing}
    \end{subfigure}
    \begin{subfigure}{0.95\linewidth}
        \includegraphics[width=\linewidth]{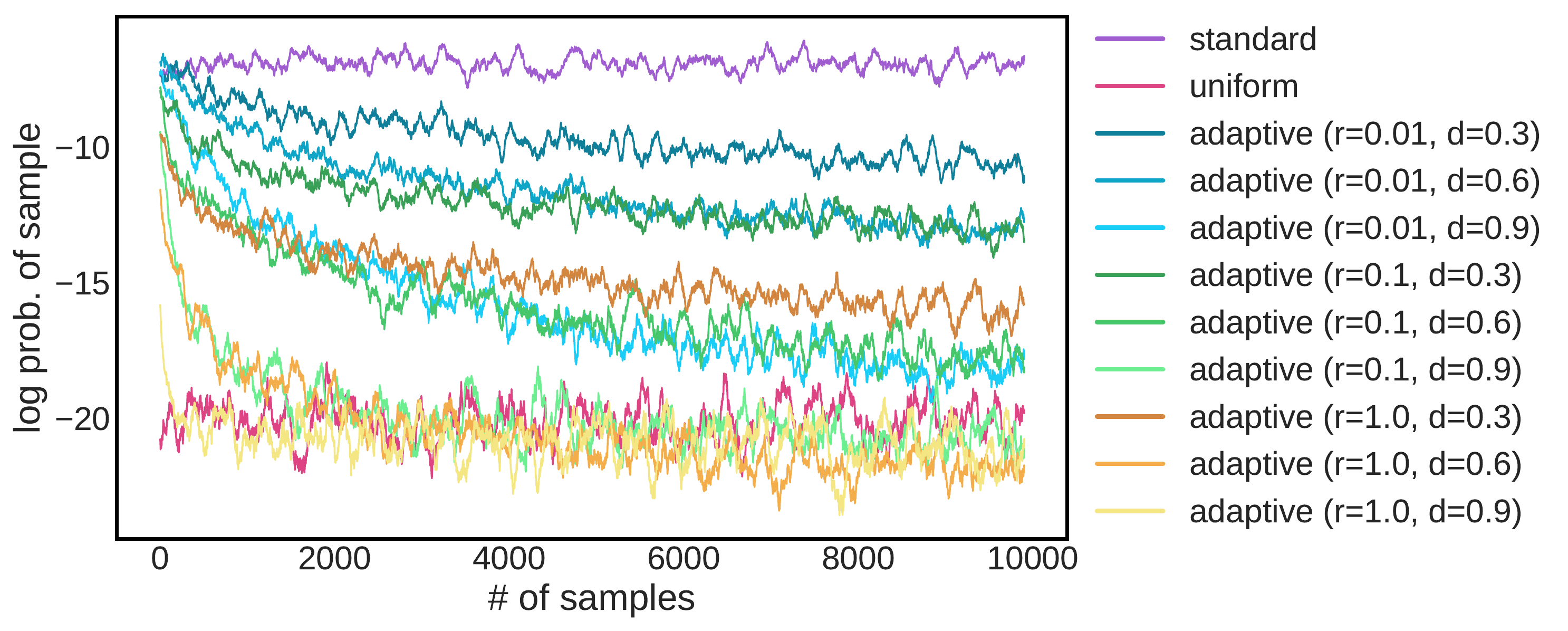}
        \caption{Likelihood of samples over time}
    \end{subfigure}
    \caption{\small Efficiency of sampling strategies for the Liftoff simulator. (a) Number of unique samples vs total samples so far. (b) Probability of sampling a unique next program given samples so far. (c) Likelihood of generated samples over time for different sampling strategies.}
    \label{fig:sampling_analysis}
\end{figure}

Fig.~\ref{fig:sampling_analysis} shows a comparison of the effectiveness of adaptive sampling to uniform and i.i.d. sampling.
Adaptive sampling interpolates nicely between sampling likely examples early on, as i.i.d. sampling does, to sampling unlikely examples later, as uniform-choice sampling does.
Note that adaptive sampling is customisable: as shown in \ref{fig:sampling_analysis}c, the algorithm has parameters ($r$ and $d$) that can be adjusted to control how fast we explore increasingly unlikely productions.

\subsection{Neural Approximate Parsing}

With good diverse samples available, we now aim to learn an approximation to
the posterior $p_\mathcal{G}(\tau | y)$.
We will do so by training a deep neural network to reconstruct the trajectory step-by-step.
We call this approach \textit{neural approximate parsing} with generative grading, or GG-NAP.

The challenge of inference over trajectories is a difficult one. Trajectories can vary in length and contain decision nodes with different support. To approach this, we decompose the inference task into a set of easier sub-tasks, similar to \cite{ritchie2016DAIPP,le2016inference}. The posterior distribution over a trajectory $\tau = (a_t, x_{a_t})_{t=1}^T$ given a production $y$ can be written as the product of individual posteriors over each decision node $x_{a_t}$ using the chain rule:
\begin{equation}
    \small
    p_\mathcal{G}(x_{a_1}, \ldots x_{a_T}|y) = \prod_{t=1}^{T} p_\mathcal{G}(x_{a_t}|y, \mathbf{x}_{<a_t})
\label{eqn:decompose}
\end{equation}
where $\mathbf{x}_{<a_t}$ denotes previous (possibly non-contiguous) nonterminals $(x_{a_1}, \ldots, x_{a_{t-1}})$. Eqn.~\ref{eqn:decompose} shows that we can learn each posterior $p(x_{a_t}|\mathbf{x}_{<a_t},y)$ separately. With an autoregressive model $\mathcal{M}$, we can efficiently represent the influence of previous nonterminals $\mathbf{x}_{<a_t}$ using a shared hidden representation over $T$ timesteps. Since most standard choices for $\mathcal{M}$ (e.g.~an RNN) require fixed-dimension inputs, we need to encode the solution and the history of choices into consistent vectors.

Firstly, to encode the solution $y$, we use standard machinery (e.g. CNNs for images, RNNs for text) with a fixed output dimension.  To represent the nonterminal choices with different support, we define three layers for each random variable $x_{a_t}$: (1) a one-hot embedding layer that uses the unique name $a_t$ to lexically identify the random variable, (2) a value embedding layer that maps the value of $x_{a_t}$ to a fixed dimension vector and (3) a value decoding layer that transforms the hidden output state of $\mathcal{M}$ into parameters of the posterior for the next nonterminal $x_{a_t+1}$. Thus, the input to the $\mathcal{M}$ is a fixed size, being the concatenation of the value embedding, name embedding, and production encoding.\footnote{Specific details can be found in the code.}

To train the GG-NAP, we optimise the objective,
\begin{equation}
    \small
    \mathcal{L}(\theta) = \mathbb{E}_{p_\mathcal{G}(\tau,y)}[\log p_\theta(\tau|y)] \approx \frac{1}{M}\sum_{m=1}^{N} \log p_\theta(\tau^{(m)} | y^{(m)})
    \label{eqn:objective}
\end{equation}
where $\theta$ are all trainable parameters and $p_\theta(\tau|y)$ represents the posterior distribution defined by the inference engine.
At test time, given only a production $y$, GG-NAP recursively samples $x_{a_{t}} \sim p_\theta(x_{a_{t}}|y, \mathbf{x}_{<a_t})$ for $t = 1, \ldots, T$ and uses each sample as the input to the next step in $\mathcal{M}$, as usual for sequence generation models \cite{graves_char_rnn}.

\subsection{kNN Baseline}

As a strong baseline for the parsing task, we consider a nearest neighbour classifier. We store our large dataset of samples $\mathcal{D} = \{ (\tau^{(m)}, y^{(m)}) \}_{m=1}^N$. At test time, given an input solution to parse, we can find its nearest neighbour in the samples with a linear search of $\mathcal{D}$, and return its associated trajectory. Depending on the problem, the solutions $y$ will be in a different output space (image, text) and thus the distance metric used for the nearest-neighbour search will be domain dependent. We refer to this baseline as GG-kNN. Note that GG-kNN is quite costly in memory and runtime as it needs to store and iterate through all samples in the dataset.

\section{Experiments}
\label{sec:experiments}

We test generative grading on a suite of education data sets focusing on introductory courses from online platforms and large universities.
For each dataset, we compare our approach to supervised learning, PCFGs, k-nearest neighbours, and human performance. In Sec.~\ref{sec:datasets}, we introduce the data sets, then present results in Sec.~\ref{sec:results}.

\subsection{Datasets}
\label{sec:datasets}

\begin{figure}[t!]
    \centering
    \begin{subfigure}{\linewidth}
        \centering
        \includegraphics[width=\linewidth]{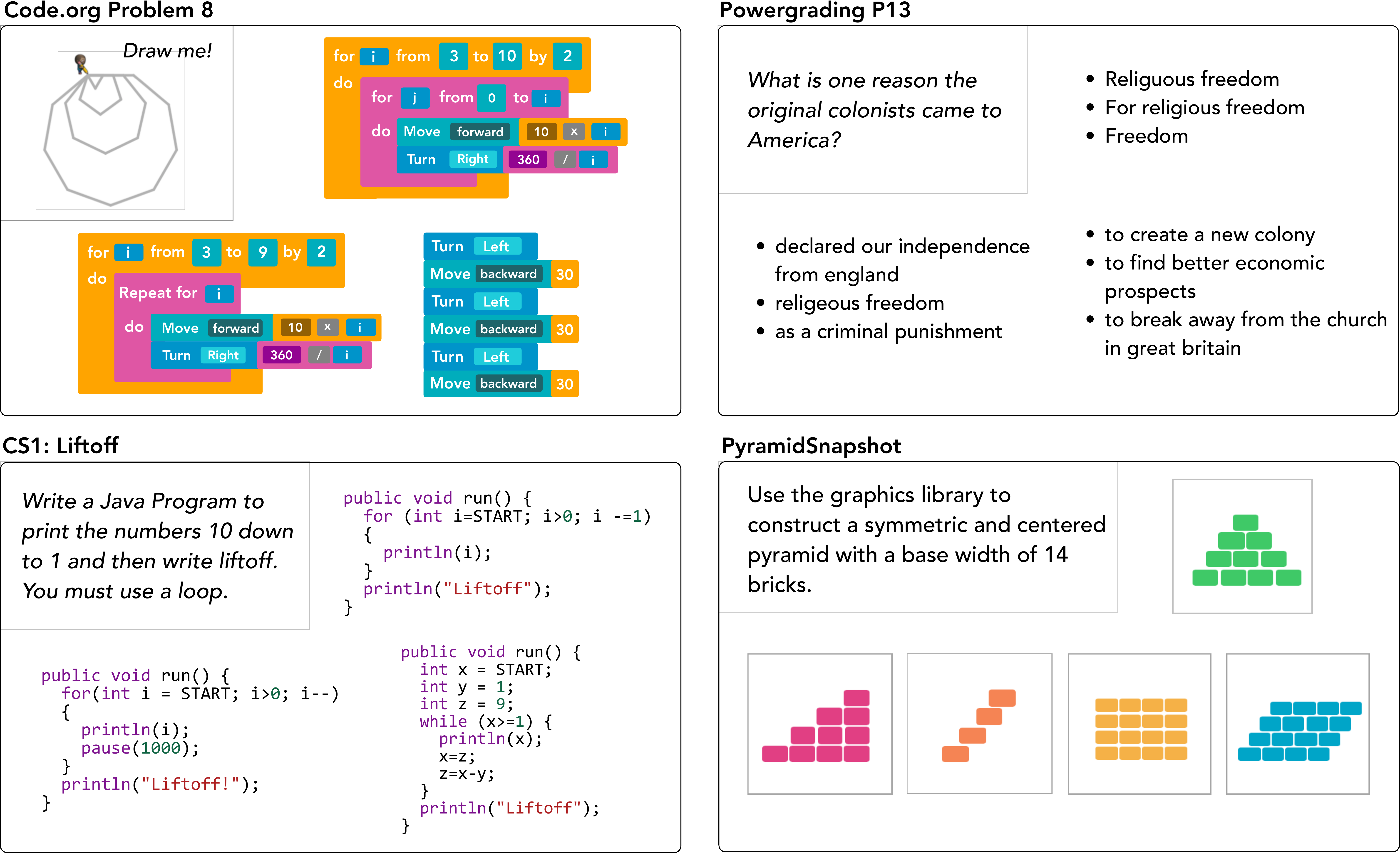}
    \end{subfigure}
    \caption{\small We show the prompt and example solutions for our four datasets.}
    \label{fig:grading_problem}
\end{figure}

We consider four educational contexts. Fig.~\ref{fig:grading_problem} shows example student solutions for each problem.

\textbf{Block-based Programming}  Code.org released a data set of student responses to eight Blocky  exercises from one of their curriculums online, which focuses on drawing shapes with nested loops. We take the last problem in the curriculum (the most difficult one): drawing polygons with an increasing number of sides---which has 302 human graded responses with 26 misconceptions regarding looping and geometry (e.g. ``missing for loop'' or ``incorrect angle'') from \cite{wu2018zero}.

\textbf{Free Response Language}  Powergrading \cite{basu2013powergrading} contains 700 responses to a United States citizenship exam, each graded for correctness by 3 humans. Responses are in natural language, but are typically short (average of 4.2 words). We focus on the most difficult question, as measured by \cite{riordan2017investigating}: ``name one reason the original colonists came to America". Correct responses span economic, political, and religious reasons.

\textbf{Graphics Programming}  PyramidSnapshot is a university CS1 course assignment intended to be a student's first exposure to variables, objects, and loops. The task is to build a pyramid using Java's ACM graphics library by placing individual blocks. The dataset is composed of \textit{images} of rendered pyramids from intermediary ``snapshots" of student work. \cite{yan_pyramid} annotated 12k unique snapshots with 5 categories representing ``knowledge stages" of understanding.

\textbf{University Programming Assignment}  Liftoff is a second assignment from a university CS1 course that tests looping. Students are tasked to write a program that prints a countdown from 10 to 1 followed by the phrase ``Liftoff''. In Sec.~\ref{sec:extensions}, we will use Liftoff for a human-in-the-loop study where experts generatively grade 176 solutions from a semester of students and measure accuracy and grading time.


\begin{figure*}[h!]
    \centering
        \includegraphics[width=0.95\linewidth]{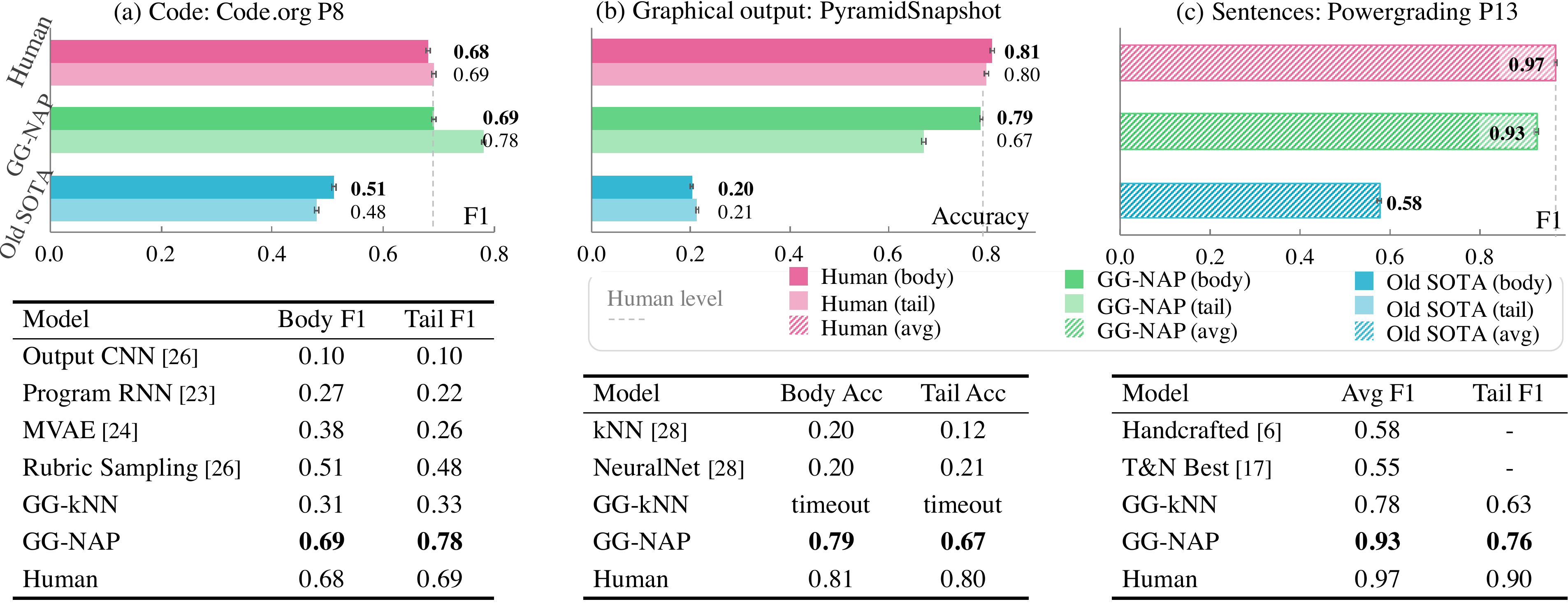}

    \caption{\small Summary of results for providing feedback to student work in three educational contexts: block-based programming, graphics programming, and free response language. Generative grading shows strong performance in all three settings, closely approximating human-level performance in two data sets, and surpassing human-level performance in the other.}
    \label{fig:results_summary}
\end{figure*}

\subsection{Simulator Descriptions}

We provide a brief overview of the Idea2Text simulators constructed for each domain.

\textbf{Block-based Programming}  The primary innovation is to use the first decision node random variable to represents student ability. This ability variable will affect the distributions for random variables later in the trajectory such as deciding the loop structure and body. The intuition this captures is that high ability students make very few to no mistakes whereas low ability students tend to make many correlated misunderstandings. This simulator contains 52 decision nodes.

\textbf{Free Response Language}  Idea2Text simulators over natural language need to explain variance in both semantic meaning and prose. We inspected the first 100 responses to gauge student thinking. Procedurally, the first random variable is choosing whether the production will be correct or incorrect. It then chooses a subject, verb, and noun dependent on the correctness. Correct answers lead to topics like religion, politics, and economics while incorrect answers are about taxation, exploration, or physical goods. Finally, we add a random variable to decide a writing style to craft a sentence. To capture variations in tense, we use a conjugator \cite{mlconjug} for the final production. This simulator contains 53 decision nodes.

\textbf{Graphics Programming}  The primary decision in this simulator decides between 13 ``strategies" (e.g. making a parallelogram, right triangle, a brick wall, etc.) that the instructor believed students would use. Each of the 13 options leads to its own set of nodes that are responsible for deciding shape, location, and colour. The production uses Java to render an image output. This simulator contains 121 decision nodes, and required looking at 200 unlabelled student solutions in its design.

\textbf{University Programming Assignment}  To model student thinking on Liftoff, this simulator first determines whether to use a loop, and, if so, chooses between ``for'' and ``while'' loop structures. It then formulates the loop syntax, choosing a condition statement and whether to count up or count down. Finally, it chooses the syntax of the print statements. Notably, each choice is dependent on previous ones. For example, choosing an end value in a for loop is sensibly conditioned on a chosen start value.  This simulator contains 26 decision nodes.

\subsection{Results for Feedback Prediction}
\label{sec:results}
We show the results of generative grading for each of the datasets above.

For each dataset, we have access to a set of real student solutions and corresponding human-provided feedback labels, which we use for evaluation.

We ask instructors to create an Idea2Text simulator for each dataset, and train the deep inference network GG-NAP using simulated student solutions. At test time, we pass a real student solution into the inference model, and get back a trajectory of the simulator. This trajectory contains decision node choices that correspond to the human-provided feedback labels, and we use these as the model's feedback prediction.

Our performance metric for evaluating the model's predicted feedback labels is accuracy or F1 score, depending on the convention of prior work. Computing an average of the metric across the evaluation dataset would over-prioritise examples that appear frequently; this is particularly important to avoid for the Zipf distributed solutions. Since we care about providing feedback to struggling students in the tail of the distribution, we separately calculate performance for different ``regions'' of the Zipf.  Specifically, we define the \textit{head} as the $k$ most popular solutions, the \textit{tail} as solutions that appear only once or twice, and the \textit{body} as the rest. As solutions in the head can be trivially memorised, we focus on performance on the body and tail.

\textbf{Training Details}
We report averages over three runs; error bars are shown in Fig.~\ref{fig:results_summary}.
We use a batch size of 64, train for 20 epochs on 100k unique samples adaptively sampled from the simulator. We optimise using Adam \cite{kingma2014adam} with a learning rate of 5e-4 and weight decay of 1e-7. For PyramidSnapshot, we use VGG-11 \cite{simonyan2014very} with Xavier initialisation \cite{glorot2010understanding} as the encoder network. For other data sets, we use a Recurrent Neural Network (RNN) with 4 layers, a hidden size of 256. The deep inference network itself is an unrolled RNN: we use a gated recurrent unit with a hidden dimension of 256 and no dropout. The value and index embedding layers output a vector of dimension 32. These hyperparameters were chosen using grid search.




\textbf{Code.org}  As feedback for Code.org exercises has been studied in prior work \cite{wu2018zero}, we compare generative grading to a suite of baselines including supervised models trained to classify misconceptions from the hand-labelled dataset (Output CNN \cite{wang2017learning} + Program RNN \cite{wu2018zero}), unsupervised models that learn a latent vector representation of student work (MVAE), to the k-nearest neighbours baseline GG-kNN from Sec.~\ref{sec:models}. Most relevant to our approach is the ``rubric sampling'' \cite{wu2018zero} comparison, which uses a PCFG to simulate students and generate a supervised data set to train a RNN classifier. Human accuracy is measured by comparing the feedback provided by multiple annotators to the mode.

As shown in Fig.~\ref{fig:results_summary}, generative grading is able to provide accurate feedback (historically measured as F1) beyond the level of individual human annotators, setting the new state-of-the-art.
We observe a large improvement over prior work, which perform significantly worse than human graders. Compared to rubric sampling, we find a 18\% (absolute) improvement in the body and a 30\% (absolute) improvement in the tail. This clearly demonstrates the practical importance of being context-sensitive. The global state of Idea2Text simulators allow us to easily write richer generative models that are capable of better simulating real students. The potential impact of a human-level autonomous grader is large: Code.org is used by 610 million students, and our approach could save thousands of human hours for teachers by providing the same quality of feedback at scale.



\textbf{Powergrading}  We find similarly strong performance on the Powergrading corpus of short answer responses to a citizenship question. Fig.~\ref{fig:results_summary} shows that generative grading reaches a F1 score of 0.93, an increase of 0.35 points above prior work that used hand-crafted features to predict correctness \cite{daxenberger2014dkpro}, and 0.38 points above supervised neural networks \cite{riordan2017investigating}. We were unable to compare to rubric sampling \cite{wu2018beyond} as it was too difficult to write a faithful PCFG to describe free response language. Generative grading takes a large step towards closing the gap to human performance, measured to be F1 = 0.97 (within 0.04). We are especially optimistic about these results as Powergrading responses contain natural language, this is promising signal that ideas from generative grading could generalise beyond computer science education.

\textbf{PyramidSnapshot} Investigating a third modality of image output from a graphics assignment, we find similar results comparing generative grading to the k-nearest neighbour baseline and a VGG image classifier presented in \cite{yan_pyramid}, outperforming the latter by nearly 50\% absolute.

Unlike other datasets, the PyramidSnapshot dataset includes student's intermediary work, showing stages of progression through multiple attempts at solving the problem. With our near-human level performance, instructors could use GG-NAP to measure student cognitive understanding over time \textit{as} students work. This builds in a real-time feedback loop between the student and teacher that enables a quick and accurate way of assessing teaching quality and characterising both individual and classroom learning progress. From a technical perspective, since PyramidSnapshot only includes rendered images (and not student code), generative grading was responsible for parsing student solutions from just images alone, a feat not possible without the flexibility of probabilistic programs used in Idea2Text. For this reason, we could not apply rubric sampling in this context either.


\begin{figure}[t!]
    \centering
    \includegraphics[width=0.85\linewidth]{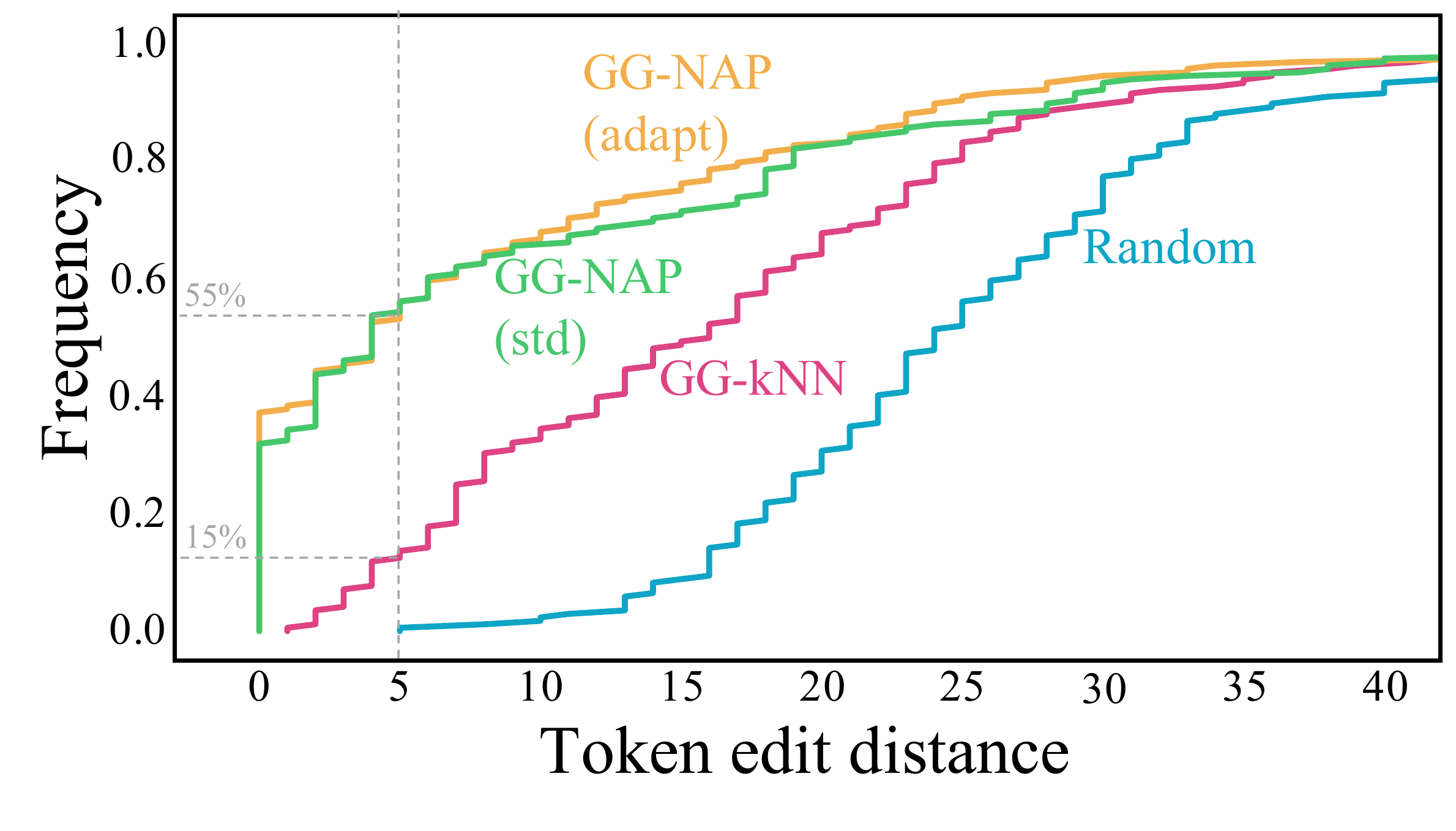}
    \caption{\small CDF of Levenshtein edit distance between student programs and nearest-neighbours using various algorithms.}
    \label{fig:liftoff_edit_dist}
\end{figure}

\section{Extensions}
\label{sec:extensions}

Our results show that generative grading is a powerful tool for the feedback prediction task. However, our system is much more general than this and has many interesting extensions that we discuss in this section.

\subsection{Nearest In-Simulator Neighbour}
\label{sec:nearest_neighbour}

As described in Section \ref{sec:inference}, our inference model learns to map a given solution, $y$, to a decision trajectory in the simulator. So far, we have used this only to provide feedback with a fixed set of labels. However, the decision trajectory has a much more powerful interpretation; it represents the sequence of decisions in the simulator that the inference model thinks will produce $y$. Since we have the simulator in hand, we can actually execute it with these predicted sequence of decisions and inspect the simulated output, $\widehat{y}$.

If the simulated output is exactly equal to the original input solution, i.e. if $y = \widehat{y}$, then we can make a strong claim: the predicted trajectory from the inference model was provably correct and the corresponding labels can be assigned with $100\%$ confidence. This is a claim that is seldom possible with traditional supervised learning methods and advances efforts towards creating explainable AI.

What about when $y \neq \widehat{y}$? In this case, the simulated output is not an exact match to the student solution, but we can still treat it as a ``\textit{nearest in-simulator neighbour}" to $y$. Fig. \ref{fig:liftoff_edit_dist} shows the quality of these nearest neighbours to the student solutions using a distance metric like edit distance.

As we show below, these nearest neighbours can be used for powerful forms of feedback mechanisms.

\subsection{Human-in-the-loop Grading}

In a real-world setting, predicting feedback labels could be unreliable due to the high risk of giving students incorrect feedback. Beyond automated feedback, we explore how generative grading can be used to make human graders more effective using a human-in-the-loop approach.

\begin{figure}[t!]
    \centering
    \includegraphics[width=\linewidth]{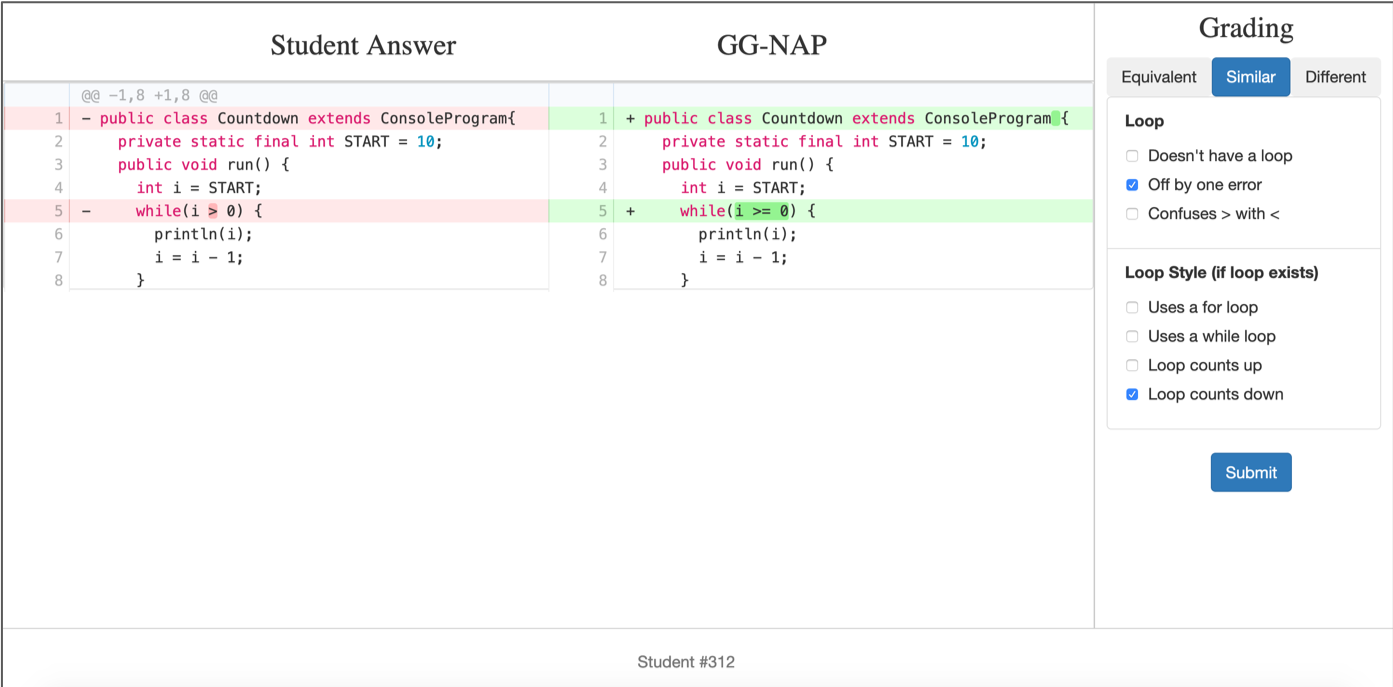}
    \caption{Human-in-the-loop Generative Grading UI}
    \label{fig:grading_ui_full}
\end{figure}

To do this, we created a human-in-the-loop grading system using GG-NAP. For each student solution, we use the inference model to find the nearest in-simulator neighbour (Sec.~\ref{sec:nearest_neighbour}); this nearest neighbour already has associated labels that are correct \textit{for the nearest neighbour}. A human grader is presented with the original student solution, as well as a diff to the nearest neighbour; the grader then adjusts the labels of the nearest neighbour based on the diff to determine grades for the real solution. We show an image of the user-interface of this system in Fig.~\ref{fig:grading_ui_full}.

We investigated the impact of this human-in-the-loop system on grading accuracy and speed in a real classroom setting. We hired a cohort of expert graders (teaching assistants with similar experience from a large private university) who graded 30 real student solutions to Liftoff.
For control, half the graders proceeded traditionally, assigning a set of feedback labels by just inspecting the student solutions.  The other half of graders additionally had access to (1) the feedback assigned to the nearest neighbour by GG-NAP and (2) a code differential between the student program and the nearest neighbour. Some example feedback labels included ``off by one increment", ``uses while loop", or ``confused $>$ with $<$". All grading was done on a web application that kept track of the time taken to grade a problem.

\begin{figure*}[t!]
    \centering
    \begin{subfigure}{0.32\textwidth}
        \centering
        \includegraphics[width=0.9\linewidth]{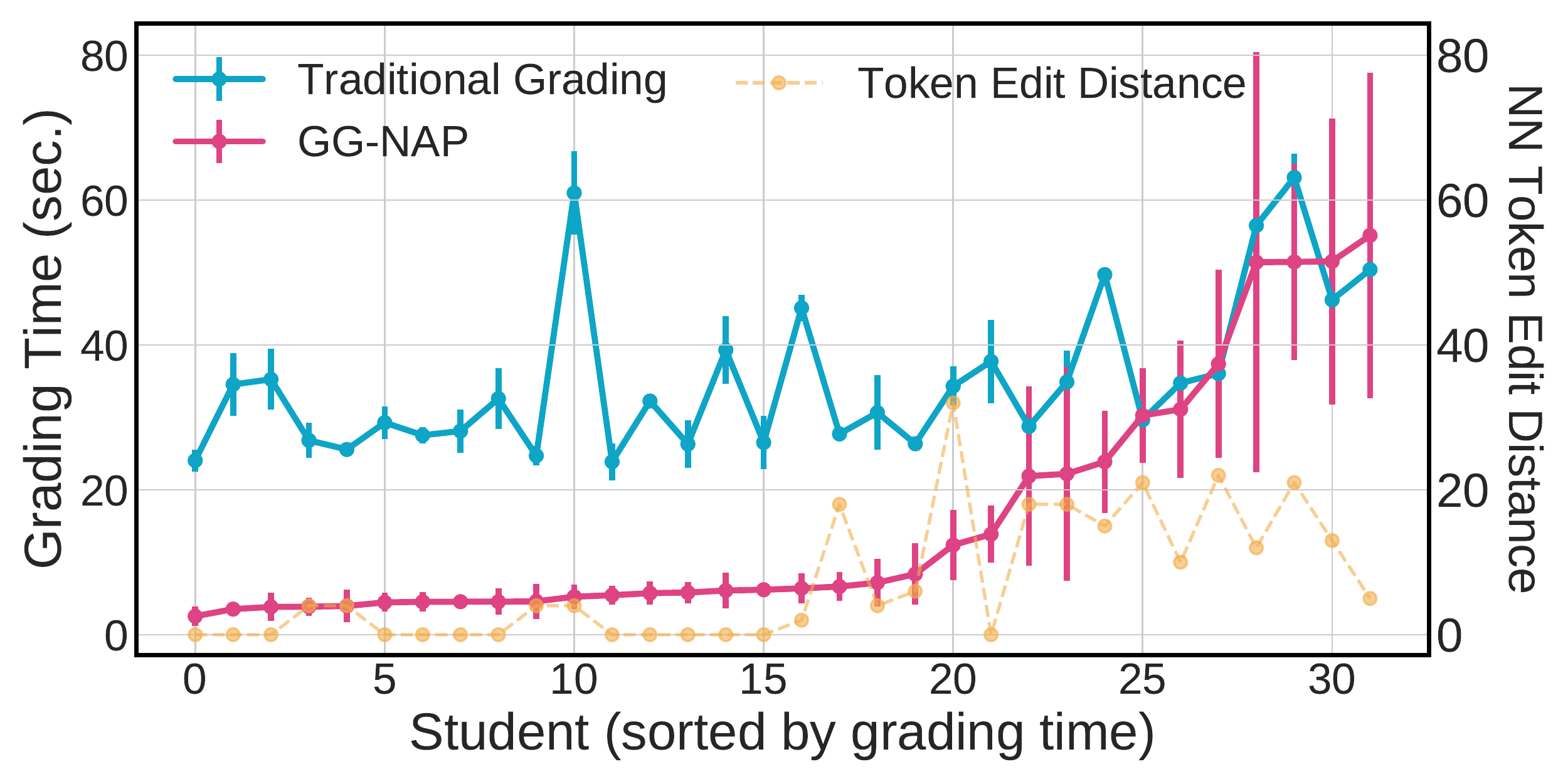}
        \caption{Classroom Experiment Results}
        \label{fig:human_exp}
    \end{subfigure}
    \begin{subfigure}{0.32\textwidth}
        \centering
        \includegraphics[width=\linewidth]{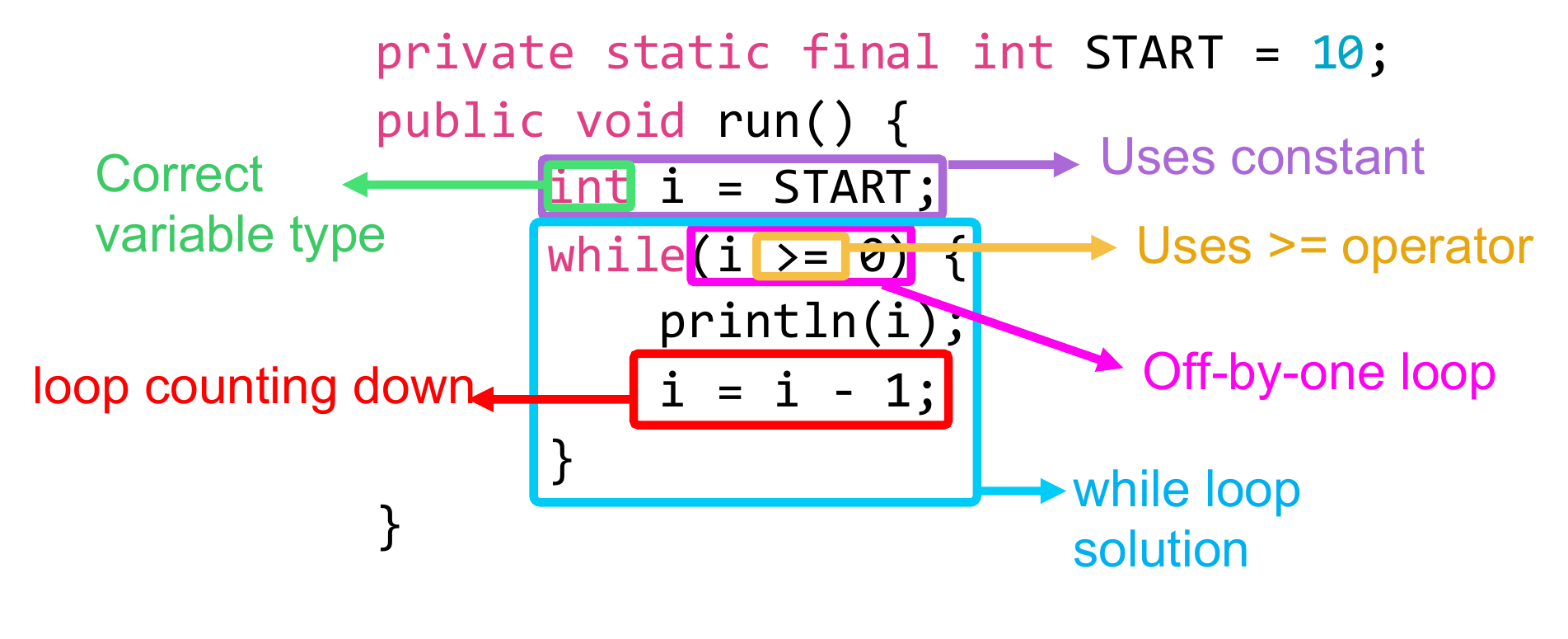}
        \caption{Automated Dense Feedback}
        \label{fig:code_highlight}
    \end{subfigure}
    \begin{subfigure}{0.32\textwidth}
        \centering
        \includegraphics[width=0.9\linewidth]{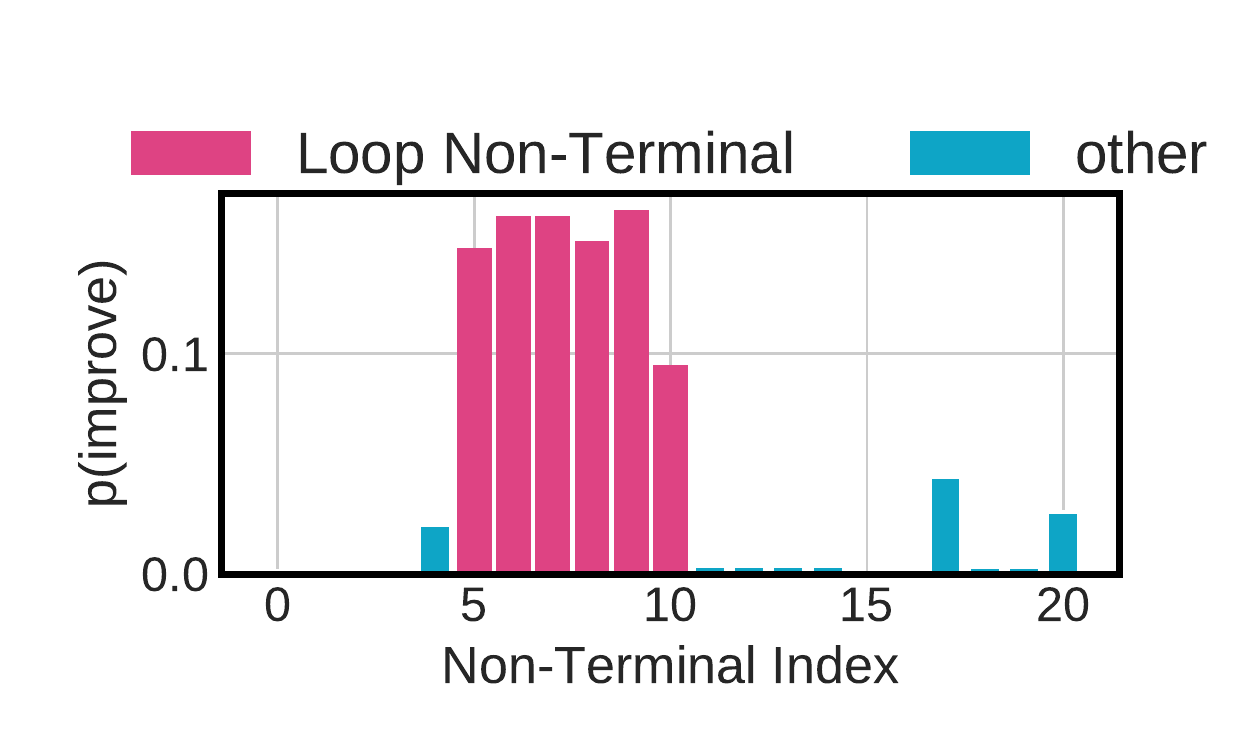}
        \caption{Auto-improving Simulators}
        \label{fig:make_better}
    \end{subfigure}

    \caption{\small (a) Plot of average time taken to grade 30 student solutions to Liftoff. Generative grading reduces grading time for 26 out of 30 solutions. The amount of time saved correlates with the token edit distance (yellow) to the nearest neighbour in the simulator. (b) Our approach allows for automatically highlighting which part of the student solution is responsible for a predicted misconception. (c) Given a Liftoff simulator that is missing a key ``decrement loop'' decision, we can automatically find decision nodes where inference often fails on real student solutions.  The highest scoring decision nodes are all correctly related looping.}
\end{figure*}

We found that the average time for graders using our system was \textit{507 seconds} while the average time using traditional grading was \textit{1130 seconds}, a more than double increase. Moreover, with our system, only \textit{3 grading errors} (out of 30) were made with respect to gold-standard feedback given by the course Professor, compared to the \textit{8 errors} made with traditional grading. Fig.~\ref{fig:human_exp} shows these results for each of the $30$ solutions.

The improved performance stems from the semantically meaningful nearest neighbours provided by GG-NAP. Having access to graded nearest neighbours helps increase grader efficiency and reliability by allowing them to focus on only ``grading the diff'' between the real solution and the nearest neighbour. By halving both the number of errors and the amount of time, GG-NAP can have a large impact in classrooms today, saving instructors and teaching assistants unnecessary hours and worry over grading assignments.

\subsection{Highlighting feedback in student solutions.}
\label{sec:highlight}

The inferred decision trajectory for a student solution can also be used to provide ``dense" feedback that highlights the section of the code or text responsible for each misunderstanding. This would be much more effective for student learning than vague error messages currently found on most online education platforms.

To achieve this, we leverage the fact that each decision node in the simulator gets recursively expanded to produce the final solution. This means it is easy to track the portions of the output that each decision node is responsible for. For decision nodes related to student confusions, we can highlight the portion of the output in the student solution which corresponds to this confusion.
Fig.~\ref{fig:code_highlight} shows a random program with automated, segment-specific feedback given by GG-NAP. This level of explainability is sorely needed in both education and AI.


\subsection{Automatically Improving Simulators}
\label{subsec:autoimprove}
Building Idea2Text simulators is an iterative process; a user wishing to improve their simulator would want a sense of where it is lacking. Fortunately, given a set of difficult examples where GG-NAP does poorly, we can deduce the decision nodes in the simulator that consistently lead to mistakes and use these to suggest components to improve.

To do this, for each nearest neighbour to a student solution we can find decision nodes that cause substring mismatches in the student solution, using regular expressions. This is possible because each decision node is responsible for a scoped substring in the nearest neighbour output solution (Sec. \ref{sec:highlight}). By finding the decision nodes where the substring often differs between the neighbour and the solution, we can identify decisions that often causes mismatches.

To illustrate this, we took the Liftoff simulator, which contains a crucial decision node that decides between incrementing up or down in a ``for" loop, and removed the option of incrementing down. We trained GG-NAP on this smaller simulator, and used a scoring mechanism to identify relevant decision nodes responsible for failing to parse student solutions that ``increment down". Fig.~\ref{fig:make_better} shows the distribution over which nodes GG-NAP believes to be responsible for the failed parses. The top 6 decisions that GG-NAP picked out all rightfully relate to looping and increments.

\section{Limitations and Future Work}
\label{sec:discussion}

\textbf{Cost of writing good simulators.} One of the most critical steps in our approach is the ability to write good Idea2Text simulators. Writing a good simulator does not require special expertise and can be undertaken by a novice in a short time. For instance, the PyramidSnapshot simulator that sets the new state of the art  was written by a first-year undergraduate within a day. Furthermore, many aspects of simulators are re-usable: similar problems will share nonterminals and some invariances (e.g. the nonterminals that capture different ways of writing \texttt{for} loops are the same everywhere). This means every additional grammar is easier to write since it likely shares a lot in structure with existing grammars. Moreover, compared to weeks spent hand-labelling data, the cost of writing a grammar is orders of magnitude cheaper and leads to much better performance.

That being said, we believe there is room for interesting future work that explores how to make grammars easy to write and improve, with the extension in Sec.~\ref{subsec:autoimprove} already make some headway in this direction. There is also room for better formalising which types of problem domains can be faithfully modelled with Idea2Text simulators, and which domains are infeasible, like general essay writing. Lastly, more sophisticated inference approaches could be explored for handling semantic invariances in student output such as code reordering or variable renaming.

\textbf{Connections to IRT.} We find an interesting parallel of our work to Item Response Theory (IRT). IRT is essentially an extremely simple generative model that relates a student parameter $\theta$ to the probability of getting a question correct or incorrect. Some of our Idea2Text simulators also incorporate a student ability parameter $\theta$ to dictate likelihoods of making mistakes at different decisions, and can thus be seen as a more expressive and nuanced extension of the IRT generative model. Exploring this further is an interesting direction of research.

\textbf{Generating questions with Idea2Text.}
We use Idea2Text simulators to model student decision-making and corresponding example solutions. This could be used to automatically generate example solutions with known issues to show students for pedagogical purposes. The Idea2Text library can also been used to generating questions corresponding to confusions instead of solutions corresponding to confusions.

\section{Conclusion}
We proposed a method for providing automated student feedback that showed promising results across multiple modalities and domains.
Our proposed feedback system is capable of predicting student decisions corresponding to a given solution, allowing us to do nuanced forms of automated feedback. With it, ``generative grading'' can be used to automate feedback, visualise student approaches for instructors, and make grading easier, faster, and more consistent. Although more work needs to be done on making powerful grammars easier to write, we believe this is an exciting direction for the future of education and a step towards combining machine learning and human-centred artificial intelligence.

%
\bibliographystyle{abbrv}
\bibliography{main}  

\balancecolumns
\end{document}